%% file: main.tex
\documentclass[10pt,twocolumn,letterpaper]{article}

\usepackage[pagenumbers]{iccv} 

\usepackage{graphicx}
\usepackage{amsmath}
\usepackage{amssymb}
\usepackage{booktabs, comment}

\usepackage{url}
\usepackage[utf8]{inputenc} 
\usepackage[T1]{fontenc}    
\usepackage{subcaption}
\usepackage[normalem]{ulem} 
\usepackage[export]{adjustbox}
\usepackage{amsfonts}       
\usepackage{nicefrac}       
\usepackage{microtype}      
\usepackage{gensymb}
\usepackage{tabularx}
\usepackage{siunitx}
\usepackage{enumitem}
\usepackage{lineno}
\usepackage{pifont}
\usepackage{upgreek}
\usepackage{bbm}
\usepackage{scalerel}
\usepackage{mathrsfs}

\usepackage[font=small,labelfont=bf]{caption}

\usepackage{stfloats}

\newcommand{\RN}[1]{%
  \textup{\uppercase\expandafter{\romannumeral#1}}%
}

\newcommand*{\rom}[1]{\expandafter\@slowromancap\romannumeral #1@}


\definecolor{navy}{rgb}{0.7, 0.1, 0.7}
\definecolor{burgundy}{RGB}{144,0,32}
\definecolor{green2}{RGB}{0,160,0}

\newcommand{\KGnote}[1]{{\color{magenta}{\bf KG: }#1}} 
\newcommand{\KG}[1]{{\color{blue}#1}} 

\newcommand{\ZAnote}[1]{{\color{teal}{\bf ZA: }#1}} 

\newcommand{\TNnote}[1]{{\color{ForestGreen}{\bf TN: }#1}} 

\newcommand{\SMnote}[1]{{\color{brown}{\bf SM: }#1}} 
\newcommand{\SM}[1]{{\color{burgundy}#1}} 

\renewcommand{\KG}[1]{{\color{black}#1}} 
\renewcommand{\SM}[1]{{\color{black}#1}} 

\renewcommand{\KGnote}[1]{{\color{magenta}}} 
\renewcommand{\ZAnote}[1]{{\color{teal}}} 
\renewcommand{\TNnote}[1]{{\color{ForestGreen}}} 
\renewcommand{\SMnote}[1]{{\color{red}}} 

\definecolor{iccvblue}{rgb}{0.21,0.49,0.74}
\usepackage[pagebackref,breaklinks,colorlinks,allcolors=iccvblue]{hyperref} 

\usepackage[capitalize]{cleveref}
\crefname{section}{Sec.}{Secs.}
\Crefname{section}{Section}{Sections}
\Crefname{table}{Table}{Tables}
\crefname{table}{Tab.}{Tabs.}

\def\paperID{xxxx} 
\def\confName{bla}
\def\confYear{2025}

\title{Switch-a-View: View Selection Learned from Unlabeled  
In-the-wild Videos} 

\author{Sagnik Majumder$^{1}$ \hspace{3mm} Tushar Nagarajan$^{1}$ \hspace{3mm} Ziad Al-Halah$^{2}$ \hspace{3mm} Kristen Grauman$^{1}$\\
$^1$UT Austin \hspace{3mm} $^2$University of Utah 
}

\begin{document}

\input{sections/abstract}

\input{sections/introduction}

\input{sections/related_work}

\input{sections/approach}

\input{sections/experiments}

\input{sections/conclusion}

{
    \small
    \bibliographystyle{ieeenat_fullname}
    \bibliography{mybib}
}

\clearpage
\input{sections/supp}


\end{document}

%% file: sections/abstract.tex
\twocolumn[{%
\renewcommand\twocolumn[1][]{#1}%
\maketitle

\begin{center}
  \centering
      \includegraphics[width=\linewidth]{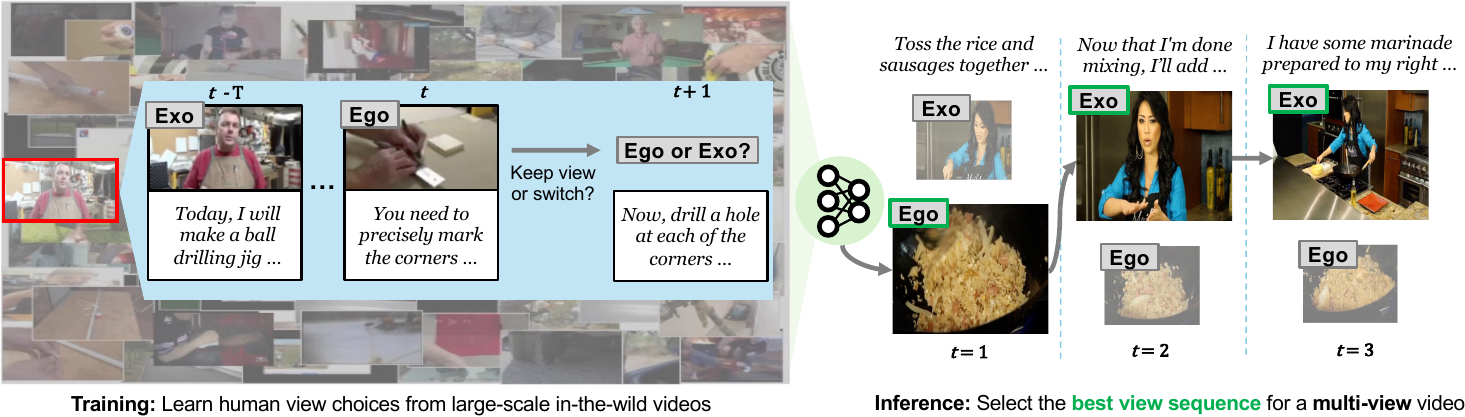}
  \captionof{figure}{
  Given a multi-view narrated how-to video, can we  select the sequence of camera viewpoints that best show the activity---automating the camerawork that is today done with manual editing?  While direct supervision for this task is impractical, our \textsc{Switch-a-view} approach shows how to learn \emph{typical viewpoint choice patterns} from large-scale unlabeled in-the-wild instructional videos (left), then translate those patterns to novel multi-view videos (right), yielding an informative how-to that hops between the most useful ego/exo viewpoints. 
  }
  \label{fig:intro}
\end{center}%
}]

\begin{abstract}
We introduce \textsc{Switch-a-view}, a model that learns to automatically select the viewpoint to display at each timepoint when creating a how-to video.  The key insight of our approach is how to train such a model from unlabeled---but human-edited---video samples.  We pose a pretext task that pseudo-labels segments in the training videos for their primary viewpoint (egocentric or exocentric), and then discovers the patterns between 
the visual and spoken content in a how-to video on the one hand and its view-switch moments on the other hand.  
Armed with this predictor, our model 
can be applied to new multi-view video settings for orchestrating
which viewpoint should be displayed when,
even when such settings come with limited labels.  
We demonstrate our idea on a variety of 
real-world video\SM{s} 
from HowTo100M and Ego-Exo4D\SM{,} and rigorously validate its advantages. Project: \url{https://vision.cs.utexas.edu/projects/switch_a_view/}.
\end{abstract}

%% file: sections/introduction.tex
\section{Introduction}\label{sec:introduction}

Video is an amazing medium for communication, and today's widely used Internet platforms make it easy to create and share content broadly.  Instructional or ``how-to" video is particularly compelling in this setting: YouTube, TikTok, and similar sites have democratized the ability to share our talents with others, by both showing and telling how to perform some special skill.  From how to plant a garden, how to make yogurt, how to fold origami, or how to give a dog a haircut, there is no shortage of how-to nuggets produced and consumed by users of many ages and backgrounds.

Creating an effective how-to video, however, is not trivial.  From potentially hours of footage from multiple cameras capturing all aspects of the instructional activity, a creator needs to edit down to the essential steps of their demonstration \emph{and} decide on the camera viewpoint (view) for each temporal segment that best reveals what they want to show.  For example, when showing how to cut the dog's hair, the instructor might first appear standing beside the dog---the camera more distant---then the camera may zoom close up to her using scissors and describing how to trim near the ear, then zoom back out while she shows progress across the dog's body.   
How-to videos
often exhibit this sequential mix of "exocentric" and "egocentric-like" viewpoints to effectively recap the procedure with clear visuals.  

The status quo is to either orchestrate camerawork live while filming, or do post-recording editing among the multiple available cameras---both of which are labor intensive. Work in automatic cinematography~\cite{10.1145/354384.354537,871033,10.1145/1324287.1324293,su2016pano2vid, hu2017deep, chen2019learning}, though inspiring, relies on heuristics or domain-specific models that are not equipped to address automatic editing of video demonstrations. 
How could we train an ``AI how-to cameraman", which, given a stream of 2 or more simultaneous camera views, could hop between them intelligently?  

Supervising this learning task presents a problem.  There are vast amounts of positive examples of well-edited how-to videos, but those edited results hide the ``negatives"---the viewpoints that were \emph{not} chosen for inclusion in the final video at any given time point.  Those are left on the cutting room floor.  This makes it unclear how to translate the editing patterns in in-the-wild edited video to new data.

To tackle this learning challenge, we design a pretext task for learning human view preferences from varying-view instructional videos on the Web.  
``Varying-view" means
that the source training videos display an arbitrary number of 
view switches 
over the course of the video (e.g., from ego to exo and back as in our example above), and contain only \emph{one} viewpoint at any time.  We introduce a model called \textsc{Switch-a-view} that learns from such data; it uses past frames together with the  how-to narrations spoken by the demonstrator to learn a binary classifier indicating whether the viewpoint is going to switch or not at the current time step.  
Then, we deploy this pretext-trained model in \emph{multi-view}, narrated video settings with limited best
view labels, and decide how to orchestrate 
the view selection of such videos
over time.
In this way, our approach captures the
view-switch patterns
from widely diverse unlabeled in-the-wild videos, then translates those trends to automatically direct the camerawork in new instances.  See 
Fig.~\ref{fig:intro}.

We train and evaluate our approach on HowTo100M~\cite{miech2019howto100m}, an extensive repository of real-world how-to videos, and 
further
show generalization 
to 
multi-view  Ego-Exo4D~\cite{grauman2023ego} videos. Our data confirms that human judges exhibit substantial agreement on what constitutes a ``best view" in a how-to video, establishing that it is possible to rigorously evaluate this task.
Furthermore, our results show 
\textsc{Switch-a-view} outperforms the state-of-the-art in multi-view video view selection~\cite{majumder2024} as well as multimodal retrieval~\cite{wang2024internvideo2} and other baselines.

%% file: sections/related_work.tex
\section{Related work}\label{sec:related}

\vspace*{-0.1in}

\paragraph{Automatic cinematography.} 
In automatic cinematography, systems automate the process of creating an effective video presentation given a video scene. Standard techniques include planning and controlling camera movements, angles and transitions.  Prior works 
target
classroom environments~\cite{10.1145/354384.354537, 10.1145/1180639.1180851, 10.1145/1324287.1324293}, group activities~\cite{10.1145/2601097.2601198}, or
(pseudo-)panoramic recordings~\cite{871033, 10.1109/TMM.2005.854388, su2016pano2vid, Xiong_2018_ECCV, chou2017self, chen2019learning, 10.1145/3183794}.
Different from all of the above, we tackle view selection in multi-view \emph{instructional} scenarios. 
Moreover, we seek a lighter-weight supervision solution: whereas prior work uses supervised discriminative methods requiring large-scale best view labels~\cite{su2016pano2vid, hu2017deep, chen2019learning} or  
bootstraps view selector training using multi-view videos annotated with view-agnostic narrations~\cite{majumder2024viewpoint},  
we aim to learn view selection 
from
readily available in-the-wild  \emph{unlabeled} instructional videos.  Furthermore, our model is multimodal, integrating both the video content as well as its transcribed speech.

\vspace*{-0.1in}

\paragraph{View selection in active perception.}

More distant from our problem, work in active perception and robotics considers how agents can intelligently select their visual input stream.  This includes next 
view selection, where
an embodied agent learns to actively place a camera for tasks like
visual recognition~\cite{8367872, ammirato2017dataset, cheng2018geometry, ramakrishnan2019emergence, Du_2023_ICCV} 
and segmentation~\cite{seifi2021glimpse, seifi2020attend}.
Whereas the objective in 
such
work is to spend less agent time or compute to see sufficient content, our goal is instead to choose an informative camera view for human consumption.  In our setting, the cameras 
are 
placed at 
certain stationary locations, or worn by a 
person performing an activity.  

\vspace*{-0.1in}
\paragraph{Weak supervision from Web data.}
Large-scale instructional data from the Web has been shown to provide weak supervision for understanding instructional activities, by aligning frames~\cite{Mavroudi_2023_ICCV} and narrations~\cite{lin2022learning, Mavroudi_2023_ICCV} with their step descriptions from 
instructional Web articles
(e.g.,  WikiHow), or through modeling the temporal order and inter-dependence of  steps~\cite{ashutosh2024video, zhong2023learning, zhou2023procedure}. Unlike any of these methods, we tackle a distinct problem of weakly supervised view-switch detection in instructional videos, with the end goal of using the detector for 
view selection.

\vspace*{-0.1in}
\paragraph{Video summarization.}
Temporal
video summarization~\cite{panda2017collaborative, 10.1007/978-3-030-58589-1_16, Badamdorj_2022_CVPR, narasimhan2022tl, He_2023_CVPR} entails creating a short but informative summary of a long video by subsampling keyframes or clips from it.
While early methods 
are
largely  
unsupervised~\cite{Smith-1995-13928, 1640912, 10.1109/CVPR.2013.350, 10.1007/978-3-319-10599-4_35}, more recent works 
derive supervision from
manual labels 
~\cite{7299154, 10.1007/978-3-319-10584-0_33, NIPS2014_0eec27c4, zhang2016summary, li2018local, rochan2018video}.  Limited work explores summarization in the context of multiple input videos~\cite{panda2017collaborative, elfeki2022multi, 8546119, 7298981}.
Video summarization and viewpoint selection are two entirely distinct tasks.  Video summarization aims to downsample the video in time to the essential parts, whereas our task essentially requires downsampling the video in \emph{space} to isolate the most informative viewpoint.

%% file: sections/approach.tex
\section{Approach}\label{sec:approach}

\begin{figure*}[!bht]
    \centering
    \includegraphics[width=\linewidth]{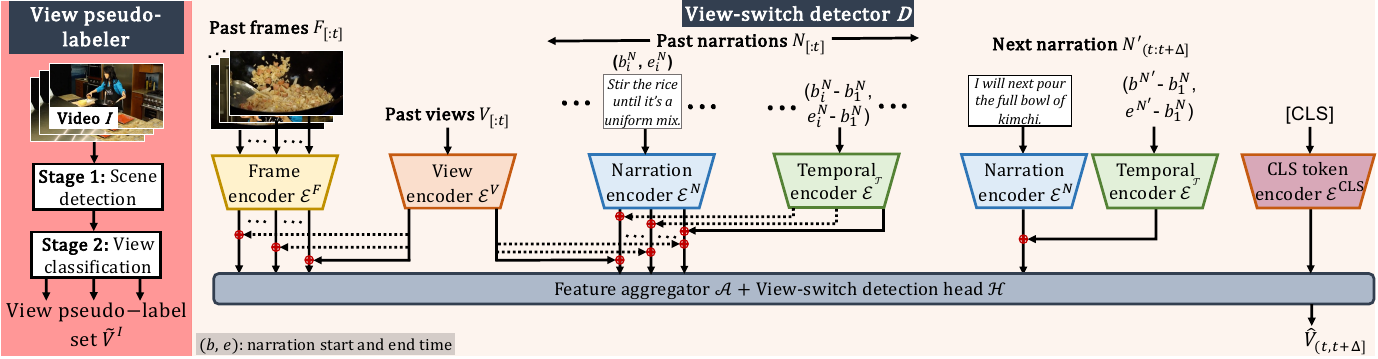}
    \caption{
    Given 
    varying-view instructional videos---videos composed of a sequence of views chosen by human(s) to accurately show the instructional activity at all 
    times---our goal is to train a view-switch detector $D$
    that 
    can predict if the view should switch or not, at any time in a new video. Our hypothesis is that such a detector, when trained on large-scale and
    in-the-wild
    videos, can capture human view preferences
    and
    facilitate 
    learning best view selection in multi-view 
    settings
    with limited labels.
    However, such in-the-wild videos lack view labels. To train nevertheless, we propose an approach comprising \textbf{(a)} a 
    view pseudo-labeler (left) that given a varying-view
    instructional video 
    $I$, 
    automatically classifies views in it and generates a pseudo-label set 
    $\tilde{V}^I$, 
    and \textbf{(b)} a view-switch detector $D$ (right) that given the pseudo-labels 
    $\tilde{V}^I$ 
    and any time $t$ in $I$, learns to predict the next view. The prediction is conditioned on the past frames, past narrations, and the next narration, where narrations are naturally occurring spoken content from the how-to demonstrator.
    }
   \vspace{-0.4cm}
    \label{fig:model}
\end{figure*}

Our goal is to train a model to predict the ``best view sequence'' for 
multi-camera 
instructional videos --- the sequence of camera viewpoints  (views) that a human would most likely select to demonstrate an instructional activity (e.g., a close-up view of ingredients in a cooking video, moving to a wide-shot view when the chef speaks and gestures).
To tackle this, we train a model for the proxy task of detecting ``view switches'' in varying-view instructional videos, which we then bootstrap to form a view selection model.

First, we formally define our pretext task (Sec.~\ref{sec:pretextTask_def}). Next, we describe how to source pseudo-labels for our pretext task by automatically classifying views in varying-view videos (Sec.~\ref{sec:pseudo_labeler}). 
We then describe our method 
and how to train it to predict view-switches (Sec.~\ref{sec:viewSwitch_detector}). 
Finally, we describe how our view-switch detector can bootstrap learning 
a 
view selection model (Sec.~\ref{sec:fewShot_viewSelection} and \ref{sec:training}) 
with limited labels.

\subsection{View-switch detection as a pretext task}
\label{sec:pretextTask_def}
We introduce our pretext task: view-switch detection in varying-view instructional videos. 
Consider a varying-view instructional video $I$, where the view changes back and forth over time between a close-up / \emph{egocentric-like} (ego) view, and a wide shot / \emph{exocentric-like} (exo) view.\footnote{We adopt this ego/exo view taxonomy given their importance and prevalence in instructional video datasets~\cite{miech2019howto100m, tang2019coin, grauman2023ego}} This results in a sequence of varying views $V$. The instructional video also contains a sequence of narrations $N$, where each narration 
$N_i$ has a start and end time, $(b_i, e_i)$, 
and provides commentary
 transcribed to text.  These narrations are free-form spoken language from the demonstrator, 
 \SM{which capture}
 their actions (``hammer the nail in there") as well as side comments (``sometimes I use my sander instead", ``thanks for watching!").

We formulate the view-switch detection task as a two-class 
\emph{view prediction} problem, where at any time $t$ in the video, the model must
detect if the view 
should be of type \emph{ego} or \emph{exo},
to best showcase the activity over the \emph{next} 
$\Delta$ seconds. 
More specifically, we require 
a model \SM{$D$} 
that predicts the human-preferred view $V_{(t, t + \Delta]}$
given the past video, narrations and views, as well as cues from the next narration. Formally,
\vspace{-0.2cm}
\begin{equation*}
\SM{D}(F_{[:t]}, N_{[:t]}, V_{[:t]}, N'_{(t, t + \Delta]}) = V_{(t, t + \Delta]}, 
\end{equation*}
where $F_{[:t]}$ is the past frames, $N_{[:t]}$ is the past narrations and $V_{[:t]}$ is the past 
views. 
$N'_{(t, t + \Delta]}$ is the next narration, if it overlaps with the prediction interval, and an empty string otherwise.
This formulation provides a path from the \emph{next-view prediction task} to the \emph{view-switch task}:
since the most recent past view is given, estimating the desired next view---and comparing it with the latest past view---is 
equivalent to predicting
whether the view switched.

While past narrations provide high-level cues about 
past activity steps
and what views were chosen to demonstrate those steps, past frames offer more fine-grained information about the same. They together form the past context that can help 
anticipate the next view. 
The next narration 
is essential to disambiguate between various potential actions that the demonstrator may do next, and the language directly hints at the appropriate views (e.g., 
the person says ``next, let's take a closer look at ..." suggesting an ego view).
Thus, combining these inputs 
will offer valuable cues to our detector.\footnote{Note that next-step narrations are also available at inference time, when we have multi-view content and a full narration track, and we aim to perform view selection.}

Critically, we aim to train this detector on large-scale and in-the-wild instructional videos~\cite{miech2019howto100m, tang2019coin}. We show that training for this pretext task can enable view selection models for multi-camera settings, with limited supervision. In short, representations developed to detect when to ``switch view'' can be repurposed with minimal modification to select the ``best view'' to switch to, since they contain rich knowledge of human-selected view-switch patterns in a large variety of in-the-wild scenarios.
Next, we show how to source pseudo-labels to train such models.

\subsection{Sourcing ``view-switch'' pseudo-labels}\label{sec:pseudo_labeler}

Instructional videos~\cite{miech2019howto100m, tang2019coin} are an ideal source of varying-view data, however they do not come paired with information about what camera viewpoint is chosen for each segment. We therefore
design a strategy to automatically identify and pseudo-label their underlying view sequences. 
We do this in two stages (Fig.~\ref{fig:model} left).

First, given a video $I$,
we use an off-the-shelf scene detector (PySceneDetect~\cite{pyscenedetect}) to compute scene boundaries. 
Using this, we split the video into a sequence of contiguous shots.  
Next, we classify each frame in the video using a pre-trained ego vs.~exo view classifier, and then aggregate the class predictions into a shot-level pseudo-label. Specifically, given a shot from $I$, we first split it into a sequence of fixed-length clips. Next, we feed each clip to the view classifier that produces the probability that the clip is from an ego vs. exo view. 
We then compute the pseudo-label for the whole shot by averaging the view probabilities across all its clips. 
We repeat these steps for all shots, and assign each frame in $I$ the same pseudo-label as the shot it lies in, to finally obtain a pseudo-label set $\tilde{V}^I$.
Combining the classifier with the scene detector reduces the overall noise in the pseudo-labels due to classification failures at scene boundaries.
We use a learned model~\cite{li2021ego} for ego-exo view classification, trained on the Charades-Ego~\cite{sigurdsson2018charades} dataset. 

\subsection{View-switch detector \SM{design}}\label{sec:viewSwitch_detector}

Given a video $I$
and any time $t$ in it, 
our view-switch detector $D$ must successfully predict the view for the future time interval $(t, t + \Delta]$. 
It must do so using the frames, narrations and views from the past, and also the next narration, if it overlaps with the prediction interval (c.f.~Sec.~\ref{sec:pretextTask_def}). See Fig.~\ref{fig:model} right.
In the following, we provide details on how our method extracts features from each input and then aggregates them for making a view prediction.

\vspace*{-0.1in}
\paragraph{Frame encoding.}
We begin by using a frame
encoder $\mathcal{E}^F$ to embed the past frames $F_{[:t]}$ and produce a visual feature sequence $f$, 
\SM{where each frame $F_i$ has a feature $f_i$.}
We further enhance each feature $f_i$ by using a view encoder $\mathcal{E}^V$ to embed the corresponding view $V^F_i$ 
into a view feature \SM{$v^F_i$}
and add it to $f_i$.
\SM{We also} encode
\SM{frame $F_i$'s}
temporal position
relative to the start time of the first past narration, \SM{denoted by $\mathcal{T}^F_i$,} using a temporal encoder $\mathcal{E}^{\mathcal{T}}$
\SM{into a feature $\tau^F_i$,}
and add \SM{it} to the \SM{enhanced} frame feature.
\SM{Formally,
\begin{equation}
f_i = \mathcal{E}^F(F_i) + \mathcal{E}^V(V^F_i) + \mathcal{E}^{\mathcal{T}}(\mathcal{T}^F_i). 
\end{equation}
}
Producing a feature per frame and augmenting it with 
view and temporal information
helps us create a \emph{fine-grained}, and 
\emph{view-} and \emph{temporally-}\emph{aware} representation that is 
anticipative of \SM{the next view}.
\vspace*{-0.1in}
  
\paragraph{Narration encoding.}
Next, we encode each past narration from $N_{[:t]}$, and the next narration $N'_{(t, t + \Delta]}$
by using an LLM encoder. This generates a text feature sequence $n$ for the past narrations and a single text feature 
$n'$ for the next narration. 

Similar to our encoding of past frames, we also make the
features for past narrations \emph{view-aware}. To do so, we first produce a per-view count of the frames that lie in the interval of each past narration $N_i$. We then estimate a dominant view $V^N_i$ for the narration---called narration view, henceforth---by setting it to the most frequent view according to the per-view frame count.    
Next, we use our view encoder $\mathcal{E}^V$ to embed the 
narration view \SM{$V^N_i$} into a view feature \SM{$v^N_i$}.
Finally, we  update the narration feature $n_i$ by adding it with 
\SM{$v^N_i$}.

Moreover, for both past and next narrations, we provide their
temporal information to our model so that it can infer the alignment between the frames and the narrations, and use it to improve its cross-modal reasoning. To this end, we first normalize the start and end time pair, $(b, e)$, 
for each \SM{past} narration \SM{$N_i$ and next narration $N'$}, 
to be relative to the start time of the first past narration. We then compute the mean time of each pair\SM{: $\mathcal{T}^N_i$ and $\mathcal{T}^{N'}$ for $N_i$ and $N'$, respectively.}
\SM{These means} convey the temporal locations of the narrations relative to each other. Next, we encode each relative mean with 
\SM{the}
temporal encoder $\mathcal{E}^\mathcal{T}$ and obtain a temporal feature\SM{: $\tau^N_i$ and $\tau^{N'}$ for $N_i$ and $N'$, respectively.}
Finally, we update 
the narration feature\SM{s} by adding them with \SM{their} temporal feature\SM{s}.
\SM{
Formally, 

\vspace{-0.2cm}
\begin{align}
    &n_i = \mathcal{E}^N(N_i)  + \mathcal{E}^V(V^N_i)  +  \mathcal{E}^{\mathcal{T}}(\mathcal{T}^N_i); \notag \\ 
    &n' = \mathcal{E}^N(N') + \mathcal{E}^{\mathcal{T}}(\mathcal{T}^{N'}).
\vspace{-0.2cm}
\end{align}
}

\paragraph{Feature aggregation and view classification.}
To aggregate the visual and narration features, we first add modality features 
$m^F$ and $m^N$ 
to the frame features 
$f$, and narration features, $n$ and $n'$,
respectively. 
\SM{Formally, 
\begin{equation}
    f_i = f_i + m^F; \quad n_i = n_i + m^N; \quad n' = n' + m^N
\end{equation}
}
These are learnable embeddings that enable our model to distinguish between the visual and text modalities, and successfully 
do cross-modal reasoning. 

We also introduce a [CLS] token in our model, and embed it with an encoder $\mathcal{E}^{\text{CLS}}$ to produce a feature 
$c$,
so that the output of our feature aggregator, which corresponds to the [CLS] token, can be used to estimate the next view. Next, we feed the frame features 
$f$, the past narration features 
$n$, the next narration feature $n'$, 
and the [CLS]-token feature $c$
into a feature aggregator $\mathcal{A}$. $\mathcal{A}$ comprises a transformer~\cite{vaswani2017attention} encoder that performs self-attention on all features and extracts multi-modal cues that are predictive of the next view. Finally, we take the output feature of $\mathcal{A}$, which corresponds to the [CLS] token, and pass it to a view classification head $\mathcal{H}$ to get an estimate $\hat{V}_{(t, t + \Delta]}$ 
of the next view 
$V_{(t, t, + \Delta]}$.
\SM{Formally,
\begin{equation}
    \hat{V}_{(t, t + \Delta]} = \mathcal{H} ( \mathcal{A}(f, n, n, c)[j_{\text{CLS}}]),
\end{equation}
where $j_{\text{CLS}}$ is the feature index for the [CLS] token.
}

\subsection{Repurposing switch detection for view selection}\label{sec:fewShot_viewSelection}
Recall that 
in view selection,
given a multi-view instructional video 
$I$
and any time $t$ in it, the goal is to predict 
the view that is preferred by
humans for showing the activity in an interval $[t, t + \Delta]$.
We introduce a view selector $S$ for tackling this task.
$S$ is a 
modification of our view-switch detector $D$,
such that
$S$ additionally has access to the \emph{frames from the simultaneously captured ego and exo views} during 
the prediction interval $[t, t + \Delta]$.

To this end, we first use our frame encoder $\mathcal{E}^F$ to embed the ego frames $F^G_{[t, t + \Delta]}$ and exo frames $F^X_{[t, t + \Delta]}$ into visual features $f^{G}$ and $f^{X}$, respectively. Next, we append $f^{G}$ and $f^{X}$ to the input sequence of our feature aggregator $\mathcal{A}$. Finally, we treat $\mathcal{A}$'s output feature for its [CLS] token input, as a representation of the best view for $[t, t + \Delta]$, and feed it to the detector's view classification head $\mathcal{H}$ to get an estimate  $\ddot{V}_{[t, t+ \Delta]}$ of the best view $V_{[t, t+ \Delta]}$.
\SM{Formally,
\begin{equation}
    \ddot{V}_{(t, t + \Delta]} = \mathcal{H} ( \mathcal{A}(f, n, n, f^G, f^X, c)[j_{\text{CLS}}]),
\end{equation}
where $j_{\text{CLS}}$ is the feature index for the [CLS] token.
}

To learn view selection 
we initialize $S$ with the parameters of our detector, trained on the view-switch detection task, and finetune it using 
a small set of
samples labeled for view selection. Next, we provide details
for training and finetuning.

\subsection{Model training objective}\label{sec:training}
We train our view-switch detector $D$ with a view classification loss $\mathcal{L}^D$. We set $\mathcal{L}^D$ to 
\begin{equation}
    \mathcal{L}^D = \mathcal{L}_{\text{CE}} (\hat{V}_{(t, t + \Delta]}, \tilde{V}_{(t, t + \Delta]}),
\end{equation}
where 
$\hat{V}_{(t, t + \Delta]}$ 
is our estimated view (c.f.~Sec.~\ref{sec:viewSwitch_detector}) and 
$\tilde{V}_{(t, t + \Delta]}$ 
is the pseudo-label from our view pseudo-labeler (c.f.~Sec.~\ref{sec:pseudo_labeler}).

To train our view selector $S$, 
we obtain a 
\emph{small}
train set of best view labels, $B$, such that 
$B = \{V_{[t_1, t_1 + \Delta]}, \ldots, V_{[t_W, t_W + \Delta]}\}$,
and $W$ is the 
label count in $B$. For each best view label 
$V_{[t_w, t_w + \Delta]} \in B$, 
and the corresponding view estimate 
$\ddot{V}_{[t_w, t_w + \Delta]}$, 
per our view selector $S$ (c.f.~Sec.~\ref{sec:fewShot_viewSelection}), we set our view selection loss $\mathcal{L}^S$ to a cross-entropy loss, such that
\begin{equation}
    \mathcal{L}^S = \mathcal{L}_{\text{CE}} (\ddot{V}_{(t_w, t_w + \Delta]}, V_{(t_w, t_w + \Delta]}).
\end{equation}

\SM{Once trained, our framework can accurately choose the preferred view in novel multi-view videos.}

%% file: sections/experiments.tex
\section{Datasets \KG{and annotations}}\label{sec:datasets}

\vspace*{-0.05in}
\paragraph{Datasets.}
We use two datasets in our experiments.
\textbf{HT100M~\cite{miech2019howto100m}} is a large-scale dataset of narrated and in-the-wild instructional videos. These videos are view-varying in nature, and the views can be broadly categorized as ego or exo. This, along with the diversity and realism of HT100M, makes it ideal for 
our view-switch detection task. 
\textbf{Ego-Exo4D~\cite{grauman2023ego}} contains multi-view videos, 
where each video is captured with five time-synced cameras---one is an ego camera worn by a human performing an instructional activity, and the other four are stationary exo cameras placed around the scene.
Moreover, the narrate-and-act (N\&A) subset of Ego-Exo4D has  videos of humans narrating and performing an activity, where the narrations are free-form and match in style
with 
HT100M,
making it compatible with our 
\SM{task of view selection with limited labels}.

\vspace*{-0.1in}

\paragraph{Training data.} To train the view-switch detector, we use 3,416 hours of HT100M videos spanning a diverse set of activities (cooking, DIY, household, etc.) and pseudo-label shots from these videos (c.f.~Sec.~\ref{sec:pseudo_labeler}). See Supp.~for details.

\vspace*{-0.1in}
\paragraph{Evaluation data.}
For evaluation, we use both HT100M and Ego-Exo4D~\cite{grauman2023ego}, where the \KG{view-switch} 
evaluation on Ego-Exo4D is \emph{zero-shot}.
While the training sets are automatically generated and pseudolabeled, we ensure a gold-standard test set free of noise by manually annotating videos for our tasks. 
\SM{To this end,} we recruit trained annotators to 
\SM{manually annotate}
the view 
\SM{types}
\SM{for} HT100M and the human-preferred views \SM{for} Ego-Exo4D.

\SM{For} HT100M, we identify 975 hours of videos
that do not overlap with our train videos above.  
We segment 4,487 fixed-length clips, each with length set 
to
the prediction interval $\Delta$ (c.f.~Sec.~\ref{sec:pretextTask_def}).
Next, we ask trained annotators to label these clips as either ego or exo.
See Supp.~for full annotation instructions and more details.

For Ego-Exo4D,
we create a  
test set containing 2.7
hours of 
N\&A
videos spanning six activity categories (cooking, bike repair, rock climbing, dancing, soccer, basketball). 
For each video, we use its ``best-exo-view" annotation from Ego-Exo4D to generate an ego-exo view pair comprising the single ego and the best exo view. 
As before, we create $\Delta$ length clips 
from each view.
We then couple the pair with its 
closest atomic activity description (time-stamped manual descriptions of the camera wearer's activity~\cite{grauman2023ego}) 
and ask our annotators to label the 
\emph{view}
between the two that 
\emph{best}
demonstrates
the 
activity described in the narration
\SM{(see Supp. Fig.~3)}.  
Importantly, this means that annotators specifically  select the ``best" view as the one that most clearly illustrates the \emph{current actions of the camera wearer}, consistent with our how-to video view selection goal.

\vspace*{-0.1in}
\paragraph{\KG{Annotator agreement on best view.}}
To ensure annotation quality for \emph{both} datasets, in addition to providing detailed annotation guidelines
and concrete examples (available in Supp.), 
we require annotators to take qualifiers with stringent passing criteria and we solicit 9 annotators' responses for each instance.
We accept an annotation only if the inter-annotator agreement is at least 78\%, meaning at least 7 out of 9 annotators agree.  This resulted in a Cohen's kappa coefficient~\cite{cohen_coefficient_1960} of 0.65 for HT100M and 0.70 for Ego-Exo4D---both of which constitute ``substantial" agreement~\cite{landis}.
This solid agreement assures the quality of our test set; despite there being some room for subjectivity in deciding the ``best" view for a how-to, this data shows human judges are indeed able to  substantially agree.

This results in a final total of 3,151 and 5,049 test instances for HT100M and Ego-Exo4D, respectively. 
In Supp.~we filter with even higher agreement thresholds, yielding even more selective (but smaller) test sets; trends for our method vs.~baselines remain \KG{consistent}.

\vspace*{-0.1in}
\paragraph{Data for view selection with limited labels.}\label{sec:fsViewSelect_dataset}
We train and evaluate our 
view selector 
on a small dataset comprising Ego-Exo4D~\cite{grauman2023ego} videos.
For our train data, we follow our annotation protocol for evaluating view-switch detection on Ego-Exo4D, and collect 
view annotations for a total of 
3.5
hours of train videos. This results in a 
total of 
6,634 train instances. For evaluation, we use our test set from view-switch detection.
\SM{This reuse is possible since 
a label indicates the type (ego/exo) of the desired next view for view-switch detection, or the desired current view for view selection.}
Train and test videos for this task are disjoint. See Supp.~for details.

\section{Experiments}\label{sec:experiments}

\paragraph{Implementation.} 
We set the durations of 
past frames
to 
8
seconds and 
past narrations
to
32
seconds, and the prediction interval to 
$\Delta = 2$ 
seconds. We set the 
sample count for view selection to $W = 5000$. We evaluate view-switch detection on HowTo100M~\cite{miech2019howto100m} by obtaining the views for the past frames (c.f.~Sec.~\ref{sec:viewSwitch_detector}) from our pseudo-labeler. For Ego-Exo4D, we adopt a teacher-forcing setup and evaluate both tasks by using the ground-truth annotations for past frames and views.
We implement our view-switch detector \SM{$D$} and view selector \SM{$S$} using the DINOv2~\cite{oquab2023dinov2} 
\SM{encoder for our frame encoder $\mathcal{E}^F$},
the Llama 2~\cite{touvron2023llama2} 
encoder for \SM{our narration encoder $\mathcal{E}^N$},  
a 8-layer transformer encoder~\cite{vaswani2017attention} 
for our feature aggregator \SM{$\mathcal{A}$}, a 2-layer MLP for the view classification head \SM{$\mathcal{H}$}, and learnable embedding layers for 
\SM{our view encoder $\mathcal{E}^V$ and temporal encoder $\mathcal{E}^{\mathcal{T}}$.}
See Supp. for more details. 

\vspace*{-0.1in}

\paragraph{Baselines.}

\begin{table}[!t]
\small
  \centering
  \setlength{\tabcolsep}{4pt}
  \resizebox{\linewidth}{!}{
    \begin{tabular}{l c c c | c c c}
    \toprule
     &  \multicolumn{3}{ c| }{HowTo100M~\citep{miech2019howto100m}} & \multicolumn{3}{ c }{Ego-Exo4D~\citep{grauman2023ego}} \\
     Model & Accuracy & AUC & AP &  Accuracy & AUC & AP\\
    \midrule
    All-ego/exo & 50.0 & 50.0 & 50.0 & 50.0 & 50.0 & 50.0\\
    Random & 52.0 & 52.0 & 51.0 & 49.3 & 49.3 & 49.7\\
    Last-frame & 42.3 & 42.3 & 53.4 & 50.0 & 50.0 & 50.0\\
    \SM{First-person pronoun detector} & 47.8 & 47.8 & 46.4 & 50.3 & 50.3 & 50.1\\
    Retrieval~\cite{wang2024internvideo2}-$F$ & \underline{53.4} & \underline{53.4} & \underline{53.2} & \textbf{52.6} & \underline{52.6} & \underline{53.6}\\
    Retrieval~\cite{wang2024internvideo2}-$N$ & 52.1 & 52.1 & 51.8 & 52.0 & 52.0 & 50.6 \\
    Retrieval~\cite{wang2024internvideo2}-$N'$ & 52.6 & 52.6 & 52.9 & 52.1 & 52.1 & 52.6\\
    \textsc{Switch-a-view} (Ours) & \textbf{59.4} & \textbf{63.8} & \textbf{60.5} & \underline{51.2} & \textbf{56.4} & \textbf{55.4}\\
    \bottomrule
  \end{tabular}
  }
  \caption{View-switch detection results. 
  Evaluation on Ego-Exo4D~\cite{grauman2023ego} is zero-shot. All values are in $\%$, and higher is better.}
  \label{tab:vsd_main}
\end{table}

\begin{table}[!t]
\small
  \centering
  \setlength{\tabcolsep}{4pt}
    \begin{tabular}{l c c c }
    \toprule
     Model & Accuracy & AUC & AP\\
    \midrule
    All-ego/exo & 50.0 & 50.0 & 50.0\\
    Random & 49.3 & 49.3 & 49.7\\
    Last-frame & 50.0 & 50.0 & 50.0\\
    \SM{First-person pronoun detector} & 50.3 & 50.3 & 50.1\\
    Retrieval~\cite{wang2024internvideo2}-$F$ & 52.3 & 52.3 & 53.6\\
    Retrieval~\cite{wang2024internvideo2}-$N$ & 51.9 & 51.9 & 51.0 \\
    Retrieval~\cite{wang2024internvideo2}-$N'$ & 52.4 & 52.4 & 52.4 \\
    View-narration~\cite{wang2024internvideo2} Similarity  & 52.5 & 52.4 & 53.9\\
    LangView~\cite{majumder2024viewpoint}-smallData & 52.1 & 52.6 & 53.2 \\   
    LangView~\cite{majumder2024viewpoint}-bigData (privileged) & \underline{53.3} & \underline{54.8} & \underline{54.5} \\   
    Ours w/o pretraining & 50.1 & 51.6 & 51.3\\
    \midrule
    \textsc{Switch-a-view} (Ours) & \textbf{54.0} & \textbf{57.3} & \textbf{56.0}\\
    \bottomrule
  \end{tabular}
  \caption{
  Results and ablation for view selection with limited labels.
  All values are in $\%$\SM{;}
  higher is better. 
  Significance $p \leq 0.05$.
  }
  \label{tab:vs_main}
  \vspace*{-0.4cm}
\end{table}

We
provide strong baselines 
comprising SOTA models and representations, as well as relevant heuristics.
For \emph{view-switch detection}, we compare against
\begin{itemize}
    \item \textbf{InternVideo2 retrieval~\cite{wang2024internvideo2}:} a set of baselines that given the most recent past frame (\textbf{Retrieval~\cite{wang2024internvideo2}-$F$}), most recent past narration (\textbf{Retrieval~\cite{wang2024internvideo2}-$N$}), or 
    \SM{next} narration 
    (\textbf{Retrieval~\cite{wang2024internvideo2}-$N'$}), first 
    encodes~\cite{wang2024internvideo2} them into fine-grained features that capture multi-frame temporal contexts, then
    uses 
    feature
    similarity 
    to retrieve a nearest neighbor of the same input type from the train set,
    and finally outputs the next view for $F$ or $N$, or the corresponding view for $N'$,
    as its prediction.\footnote{See Supp. for parallel evaluation with CLIP~\cite{radford2021learning}-style encoders, which generally underperformed InternVideo2~\cite{wang2024internvideo2} encoders.}
  \item \textbf{All-ego, All-exo, Random, Last-frame}: these are 
    heuristics
    that use the ego view (\textbf{All-ego}), the exo view (\textbf{All-exo}), a randomly chosen (\textbf{Random}) view, or the view of the most recent past frame (\textbf{Last-frame}), as their prediction.
    \item \SM{\textbf{First-person pronoun detector}: a heuristic that predicts exo when it detects first-person pronouns like ``I", ``We", ``My" or ``Our" in the next narration, as human editors often use a wide shot that reveals their face or full body, when using such pronouns.}
\end{itemize}
\vspace{0.2cm}

For \emph{view selection with limited labels}, 
in addition to the baselines listed above, we compare against the following:
\begin{itemize}
    \item \textbf{
    LangView~\cite{majumder2024viewpoint}}: a SOTA view selector that uses narrations for weakly supervised pretraining. We finetune this model with our Ego-Exo4D labels (Sec.~\ref{sec:fsViewSelect_dataset}). 
    We evaluate two versions of this baseline: LangView-\emph{bigData} and LangView-\emph{smallData}, which use large-scale Ego-Exo4D~\cite{majumder2024viewpoint} videos, and our same small subset (Sec.~\ref{sec:fsViewSelect_dataset}), respectively, for pretraining.  Note that the \emph{bigData} variant enjoys access to 
    \SM{\textbf{98x}}
    more training samples than our method, an advantage for the baseline.
    \item \textbf{View-narration~\cite{wang2024internvideo2} Similarity (VN-Sim)}: 
    separately computes the 
    cosine
    similarity between the InternVideo2 features~\cite{wang2024internvideo2} for each view and 
     the \SM{next} narration, 
    and picks the view 
    most similar to the narration.
\end{itemize}

LangView evaluates how our model fares against SOTA view selection, while the retrieval and view-narration baselines analyze whether SOTA video-language embeddings are sufficient for this task.
The heuristics verify the challenging nature of the tasks.

\vspace*{-0.1in}

\paragraph{Evaluation metrics.}

We consider three metrics: 
1) \textbf{Accuracy}, which directly measures the agreement between our predictions and labels; 2) \textbf{AUC}, the area under the ROC curve; 
and 3) \textbf{AP}, the average precision (AP) of the precision vs. recall curve.
We use AUC and AP to account for the possible class imbalance in our collected annotations. Moreover, for each metric, we separately compute its value for the \emph{same-view} and \emph{view-switch} instances in our test sets, and report the mean. This lets us account for differences in the same-view and view-switch frequency, and obtain unbiased performance measures.

\vspace{-0.2cm}
\paragraph{View-switch detection.}

In Table~\ref{tab:vsd_main}, we report our view-switch detection results.
The 
\SM{heuristics}
generally perform the worst on both datasets, underlining the challenging nature of the task. The Retrieval~\cite{wang2024internvideo2} baselines improve over them, 
indicating that our model inputs do provide cues 
about the
the view type. Among the Retrieval baselines, retrieving 
using the most recent past frame 
performs the best, showing that the past frames offer fine-grained task-relevant information beyond the narration words.
Moreover,
retrieving
with the \SM{next} narration 
is better than 
retrieving with
the most recent past narration, 
revealing that 
the \SM{next} narration
carries more pertinent details about the desired view. This is likely because
the \SM{next} narration 
is better aligned with the time interval for which the view is being predicted. 

Our method outperforms all baselines on both datasets, with 
the \SM{AUC} margin \SM{over the best baseline, Retrieval~\cite{wang2024internvideo2}-$F$,} 
being as high as 10.4\% on HowTo100M (HT100M)~\cite{miech2019howto100m} and 
3.8\%
on Ego-Exo4D~\cite{grauman2023ego}.
Our improvement over the Retrieval baselines show that 
computing feature~\cite{wang2024internvideo2}-level similarities
are not enough for this task. Instead, learning it by leveraging complementary cues from both narrations and frames is critical.
Moreover, our zero-shot results on Ego-Exo4D 
speak to our model's efficacy vis-a-vis learning human view patterns from large-scale and in-the-wild videos, which generalize to different 
scenarios, without any training.

\vspace*{-0.1in}

\paragraph{View selection.}

Table~\ref{tab:vs_main} shows our 
results on view selection with limited labels.
For the heuristics
and Retrieval~\cite{wang2024internvideo2} baselines, we observe the same performance trends as view-switch detection. 
The 
View-narration~\cite{wang2024internvideo2} Similarity (VN-Sim)
baseline 
marginally improves 
over these methods\SM{,}
indicating the frames from candidate views when combined with the corresponding narration ($N'$) provide direct cues about the preferred view.
LangView~\cite{majumder2024viewpoint}'s results benefit from its language-guided training, generally outperforming VN-Sim.

Our method significantly improves over all baselines, 
with \SM{the AUC} margin \SM{over the best baseline, LangView~\cite{majumder2024viewpoint}-bigData, being} . 
\SM{2.5}\%. 
Our gains over
VN-Sim
underscore 
that using 
feature similarity
to match the activity described in 
the \SM{next} narration 
with a candidate view does not suffice, and instead a model like ours, which can leverage multi-modal cues from the combination of both past and candidate frames, 
and past and \SM{next} narrations, 
is valuable for this task.
Training our model from scratch with only the small set of best view labels (``ours w/o pretraining") is significantly weaker, showing that our view-switch pretraining idea is doing the heavy lifting.

\begin{figure}[t]
    \centering
    \begin{subfigure}[b]{0.45\linewidth}
    \centering
    \includegraphics[width=\linewidth]{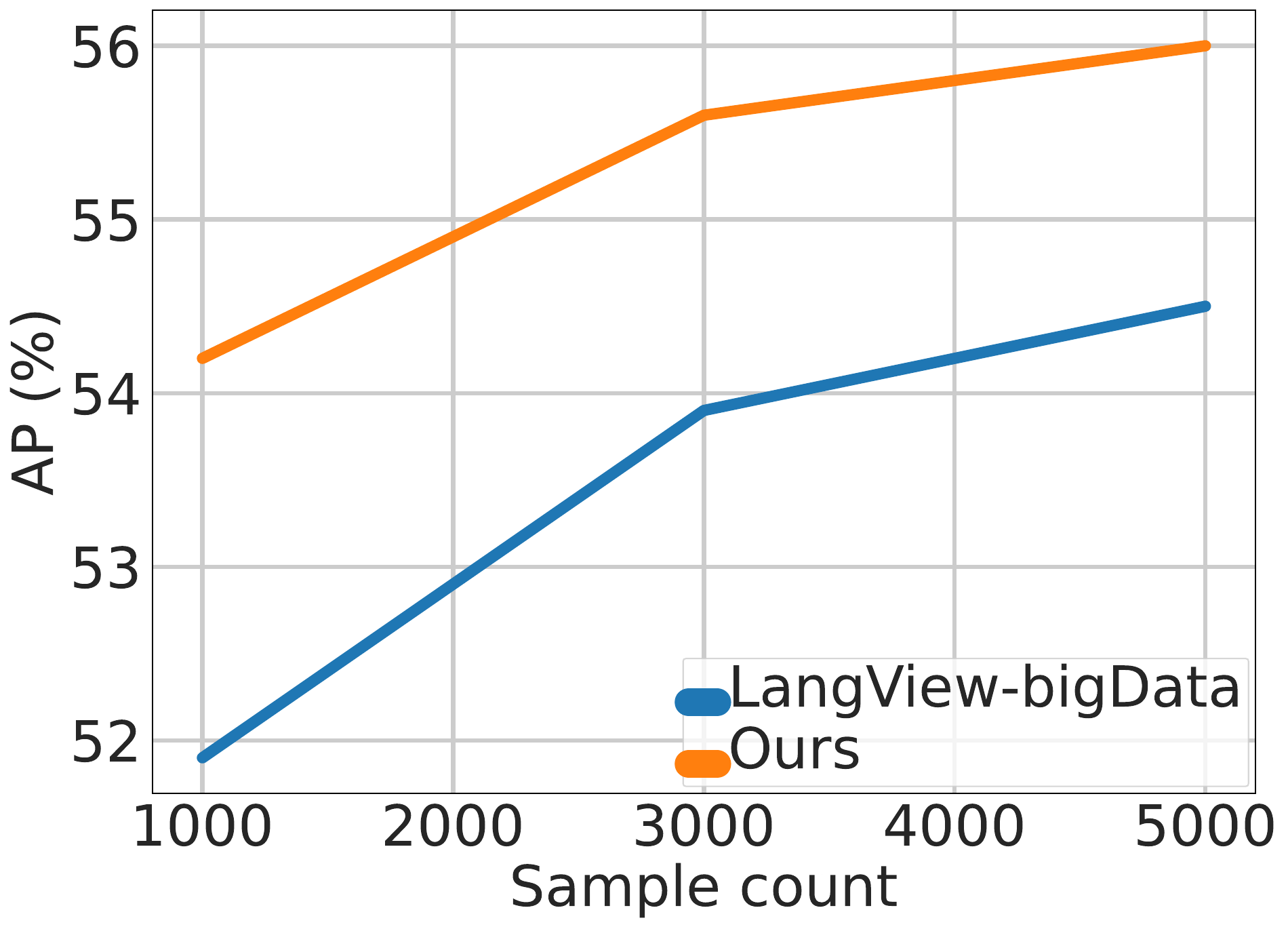}
    \caption{VS mAP vs. sample count}
    \label{fig:effectOfFewshotSampleCount}
    \end{subfigure}\hfill
    \begin{subfigure}[b]{0.45\linewidth}
    \centering
    \includegraphics[width=\linewidth]{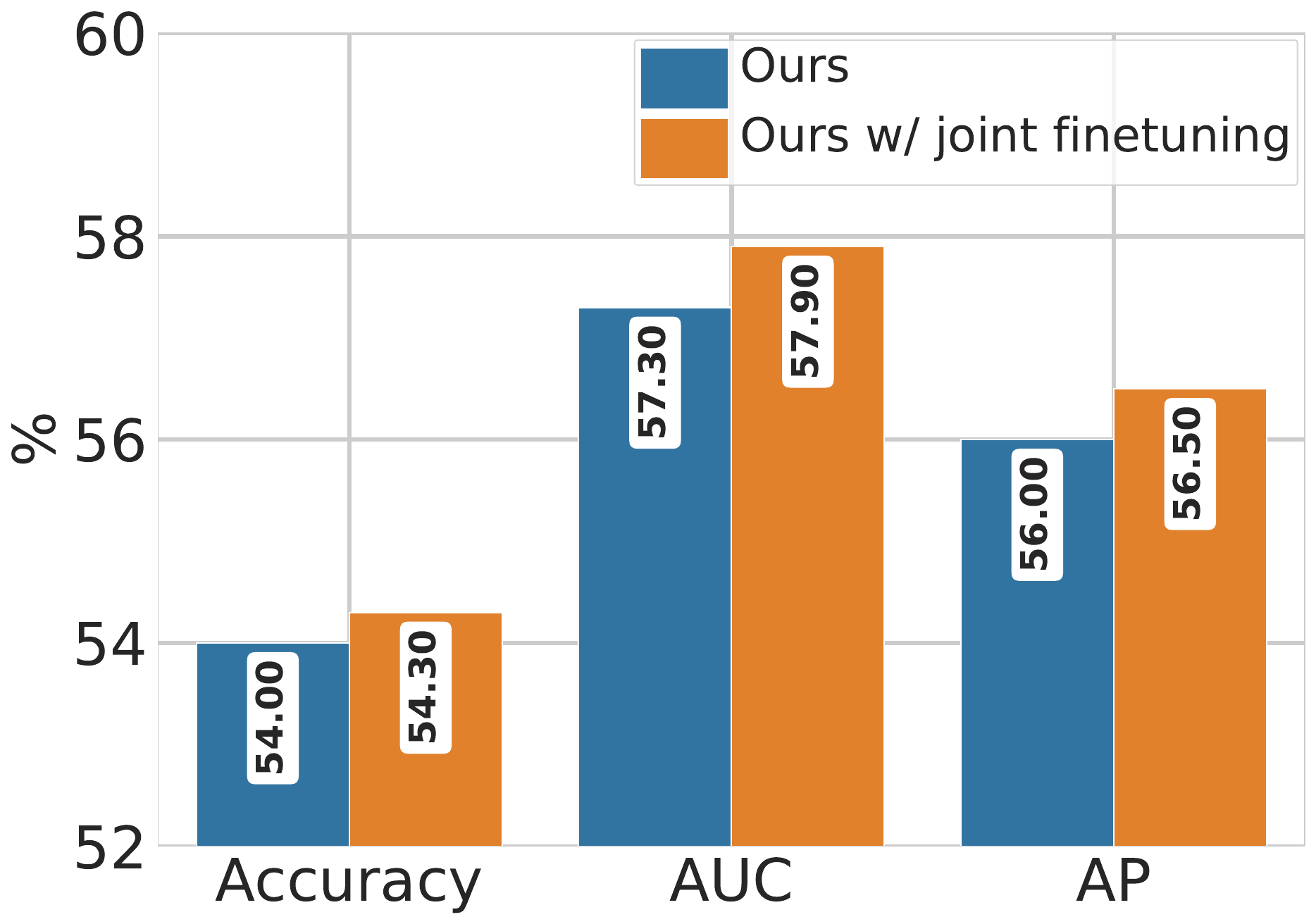} 
    \caption{VS w/ and w/o joint finetuning}
    \label{fig:ours_wNwoJointFinetune}
    \end{subfigure}
    \vspace*{-0.3cm}
\caption{(a) Effect of sample count on our view selection (VS) performance;
(b) Impact of joint finetuning with narration-based pseudo-labels~\cite{majumder2024viewpoint} and best view labels on view selection (VS) 
}\label{fig:effectOfFewshotSampleCount__n__ours_wNwoJointFinetune}
    \vspace{-0.4cm}
\end{figure}

Our gains over the SOTA LangView~\cite{majumder2024} show that learning view selection from language is less effective than that from large-scale human-edited videos, even when the videos and language are available at scale (bigData).
Moreover, the insights of LangView and our work are complementary.  We find if we fine-tune \textsc{Switch-a-view} with LangView's narration-based pseudo-labels, in \emph{addition} to our labels (Sec.~\ref{sec:fsViewSelect_dataset}), we achieve further gains.
See Fig.~\ref{fig:ours_wNwoJointFinetune}, and Supp.~for experiment details.

\vspace*{-0.15in}
\paragraph{Ablations.}

\begin{table}[t]
\small
  \centering
  \setlength{\tabcolsep}{4pt}
  \resizebox{\linewidth}{!}{
    \begin{tabular}{l c c c | c c c}
    \toprule
     &  \multicolumn{3}{ c| }{HowTo100M~\citep{miech2019howto100m}} & \multicolumn{3}{ c }{Ego-Exo4D~\citep{grauman2023ego}} \\
     Model & Accuracy & AUC & AP &  Accuracy & AUC & AP\\
     \midrule
    $N$-only & 53.5 & 54.4 & 52.3 &  50.0 & 48.7 & 49.0\\
    $N'$-only & 55.4 & 57.8 & 56.2 & 49.8 & 49.8 & 50.0\\
    $F$-only & 53.3 & 54.5 & 54.7 & 51.0 & 53.4 & 53.2\\
    $(F, N')$-only & 55.5 & 60.1 & \underline{58.1} & \textbf{52.1} & 54.2 & 52.6\\
    $(N, N')$-only & \underline{57.5} & 59.3 & 56.6 & 50.0 & 53.0 & 52.6\\
    $(F, N)$-only  & 56.0 & \underline{60.9} & 57.4 & 51.8 & \underline{54.9} & \underline{54.2}\\
    \textbf{Ours} & \textbf{59.4} & \textbf{63.8} & \textbf{60.5} & \underline{51.2} & \textbf{56.4} & \textbf{55.4}\\
    \bottomrule
  \end{tabular}
  }
  \caption{
  Ablation study for view-switch detection. All values are in $\%$, and higher is better. Significance $p \leq 0.05$.
  }
  \label{tab:vsd_ablation}
  \vspace*{-0.1cm}
\end{table}

Table~\ref{tab:vsd_ablation} shows our ablation results for view-switch detection. Dropping any one input to our model degrades performance, indicating that each input plays a role. 
Dropping two inputs hurt the performance even more, showing that higher amounts of inputs is better in any 
combination, possibly because our model design enables extracting complementary cues from them in all configurations. Moreover, using past frames instead of narrations
improves performance, re-affirming that vision provides fine-grained features necessary for high performance. Finally, using $N'$ instead of $N$ also helps improve performance in some cases,
illustrating the \SM{next} narration's 
role.

In Fig.~\ref{fig:effectOfFewshotSampleCount}, we study the effect of 
sample count 
on view selection.
Our model already improves
performance 
with as few as 1000 samples\SM{.}
This plot also highlights the low-shot success of our model vs. the 
best
baseline, 
LangView-bigData.

See Supp. for more analysis, including the effect of the past frame and narration durations on model performance, and its scenario-level breakdown.

\vspace{-0.2cm}
\paragraph{Qualitative examples.}
\begin{figure*}[!tb] 

    \centering
    \includegraphics[width=0.9\linewidth]{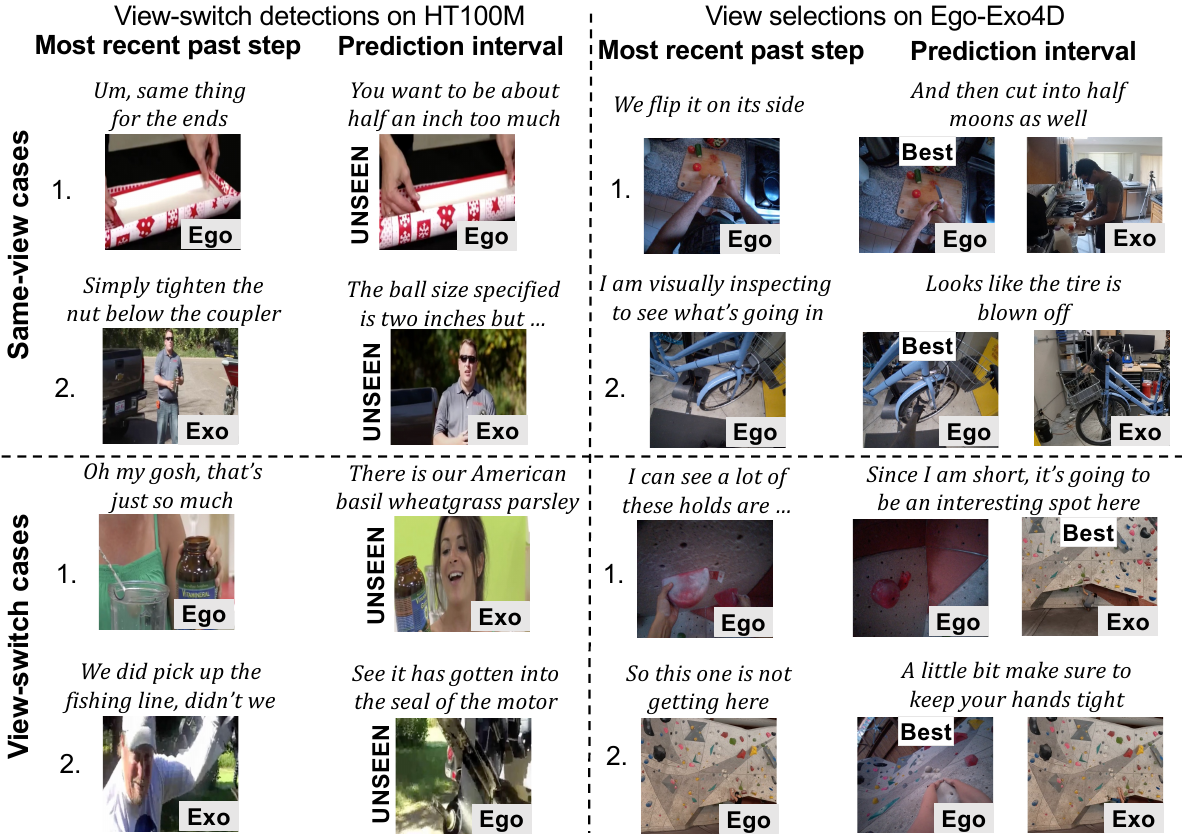}
    \caption{\textbf{Left:} successful view-switch detections by our model on same-view (\textbf{top}) and view-switch cases (\textbf{bottom}). Our model correctly detects view switches by popenopenotentially anticipating the next step using past frames (same-view sample 1, and view-switch sample 2) or leveraging the content of the next narration (same-view sample 2, and view-switch sample 1 and 2).  \textbf{Right:} successful view selections by our model on same-view (\textbf{top}) and view-switch cases (\textbf{bottom}). For view selection as well, our model can  predict the desired next view by relying on the next narration (same-view sample 1, and view-switch sample 1 and 2), or anticipate it using the past narrations (same-view sample 1 and 2), or the past frames (same-view sample 1). These examples show that all three inputs play a role in our model predictions.
    }
\label{fig:qual}
\vspace{-0.25cm}
\end{figure*}

Fig.~\ref{fig:qual}-left shows our model's successful view-switch detections on both same-view (\textbf{top}) and view-switch cases (\textbf{bottom}); see caption for details.
We also notice some common \textbf{failure modes} with our model. For view-switch detection, our model sometimes fails  when there is no next narration overlapping with the prediction interval, and neither the past frames nor narrations are predictive of the next view. In another 
failure 
type,
the past views are wrongly categorized by our pseudo-labeler for HowTo100M~\cite{miech2019howto100m} or by professional annotators for Ego-Exo4D~\cite{grauman2023ego}.
This leads to our model getting confused and predicting the wrong next view. For view selection, in addition to 
these failures, our model can fail when
both views look equally good. 
See Supp. for video examples.

%% file: sections/conclusion.tex
\section{Conclusion}

We provided an approach for learning to select views from instructional video by bootstrapping already-edited, unlabeled in-the-wild content.  Results show the
method's efficacy.
In future work, we plan to generalize to \emph{continuous} 
view selection,
potentially by integrating ideas from new view synthesis, and 
will explore 
modeling user attention for personalized view selection.


%% file: sections/supp.tex
\section{Supplementary material}
In this supplementary material we provide 
additional details about:
\begin{itemize}
    \item Video for qualitatively illustrating of our main idea and also 
    qualitatively evaluating of our view-switch detections and view selections (Sec.~\ref{sec:supp_vid}), as mentioned in 
    \SM{`Qualitative examples' in Sec.~\ref{sec:experiments}}
    in main
    \item Analysis of the impact of our shot-level pseudo-labeling on view-switch detection performance (Sec.~\ref{sec:ablations_shotLevelPseudolabeling}), as referenced in 
    \SM{Sec.~\ref{sec:pseudo_labeler}}
    in main
    \item \SM{Annotation filtering and model evaluation with higher inter-annotator agreement thresholds (Sec.~\ref{sec:supp_higherInterAnnotatorAgreementThreshold}), as noted in `Annotator agreement on best view' in Sec.~\ref{sec:datasets} in in main}
    \item \SM{View-selection results upon finetuning our view-switch detector jointly with narration-based pseudo-labels and our best view labels (Sec.~\ref{sec:weakSuperv_switchPlusLavis}), as mentioned in 
    `View selection' in Sec.~\ref{sec:experiments}
    in main}
    \item Analysis of the impact of the duration of our past frames on view-switch detection performance (Sec.~\ref{sec:supp_pastFrameDurn})\SM{, as noted in 
    `Ablations' in Sec.~\ref{sec:experiments}
    in main}
    \item Analysis of the impact of the duration of our past narrations on view-switch detection performance (Sec.~\ref{sec:supp_pastNrrnDurn})\SM{, as referenced in 
    `Ablations' in Sec.~\ref{sec:experiments}
    in main}
    \item \SM{Scenario-level breakdown of view selection performance (Sec.~\ref{sec:sceneLevel_breakdown}), as noted in 
    `Ablations' in Sec.~\ref{sec:experiments}
    in main}
    \item \SM{Feature similarity baseline evaluation with CLIP~\cite{radford2021learning}-style encoders (Sec.~\ref{sec:baseline_w_clipEnc}), as mentioned in 
    `Baselines' in Sec.~\ref{sec:experiments}
    in main}
    \item Dataset details (Sec.~\ref{sec:supp_dataset}) in addition to the ones provided in 
    \SM{
    Sec.~\ref{sec:pseudo_labeler}, and `Training data', `Evaluation data' and `Data for view selection with limited labels' in Sec.~\ref{sec:datasets}
    in main}
    \item Annotation details (Sec.~\ref{sec:supp_annotation}) in addition to the ones provided in 
    \SM{
    `Evaluation data' and `Annotator agreement on best view' in Sec.~\ref{sec:datasets}
    in main}
    \item  Additional implementation details (Sec.~\ref{sec:supp_implementation}), as referenced in 
    \SM{
    `Implementation' in Sec.~\ref{sec:experiments}
    in  main}

\end{itemize}

\subsection{Supplementary video}\label{sec:supp_vid}
The supplementary video, available at \url{https://vision.cs.utexas.edu/projects/switch_a_view/}, qualitatively illustrates our task, 
\SM{View Selection with Limited Labels,}
and our main idea towards tackling that task, Weakly-Supervised Learning from 
\SM{Unlabeled In-the-wild}
Videos.
We also show successful predictions by our model for both view-switch detection and view selection. For view selection, we additionally provide multi-step selection examples, where our model selects the best view over multiple consecutive steps. Finally, we illustrate our model's common failure modes 
\SM{(`Qualitative examples' in Sec.~\ref{sec:experiments} in main)}
with qualitative examples.

\subsection{Shot-level pseudo-labeling}\label{sec:ablations_shotLevelPseudolabeling}

\begin{table}[!t]
\small
  \centering
  \setlength{\tabcolsep}{4pt}
    \begin{tabular}{l c c c}
    \toprule
     Model & Accuracy & AUC & AP\\
     \midrule
    Ours w/o shot-level pseudo-labeling & 51.5 & 51.9 & 52.3\\
    \textbf{Ours} & \textbf{59.4} & \textbf{63.8} & \textbf{60.5}\\
    \bottomrule
  \end{tabular}
  \caption{
  Analysis of the impact of our shot-level pseudo-labeling strategy on view-switch detection performance on the HowTo100M~\cite{miech2019howto100m} dataset. All values are in $\%$, and higher is better. Significance $p \leq 0.05$.
  }
  \label{tab:supp_shotLevelPseudolabeling}
  \vspace*{-0.3cm}
\end{table}

In Table~\ref{tab:supp_shotLevelPseudolabeling}, we report the results for an additional ablation study, in which we analyze the impact of our shot-level pseudo-labeling strategy 
\SM{(Sec.~\ref{sec:pseudo_labeler} in main)}
on view-switch detection with the HowTo100M~\cite{miech2019howto100m} dataset. Upon replacing our shot-level pseudo-labeling strategy with a clip-level pseudo-labeling strategy 
\SM{(Sec~\ref{sec:pseudo_labeler} in main)}, 
we observe a drastic drop in model performance. This demonstrates that our pseudo-labeler is able to mitigate noisy clip-level predictions, particularly at scene boundaries.

\subsection{\SM{Inter-annotator agreement threshold}}\label{sec:supp_higherInterAnnotatorAgreementThreshold}
\begin{table*}[!t]
\small
  \centering
  \setlength{\tabcolsep}{4pt}
    \begin{tabular}{l c c c | c c c}
    \toprule
     &  \multicolumn{3}{ c| }{View-switch detection on HT100M} & \multicolumn{3}{ c }{View selection on Ego-Exo4D} \\
     Model & 78\% & 89\% & 100\% & 78\% & 89\% & 100\%\\
     \midrule
    Retrieval~\cite{wang2024internvideo2}-$F$ & 53.2 & 53.6 & 53.6 & \_\_ & \_\_ & \_\_  \\
    View-narration~\cite{wang2024internvideo2} Similarity & \_\_ & \_\_ & \_\_ & 53.9 & 54.2 & 53.7 \\
    LangView~\cite{majumder2024viewpoint}-bigData & \_\_ & \_\_ & \_\_ & 54.5 & 54.9 & 54.1 \\
    \textbf{Ours} & \textbf{60.5} & \textbf{60.8} & \textbf{60.7} & \textbf{56.0} & \textbf{56.2} & \textbf{55.3}\\
    \bottomrule
  \end{tabular}
  \vspace{-0.3cm}
  \caption{\SM{Model performance (AP) vs. inter-annotator agreement threshold. \_\_ indicates that the baseline is not applicable for the particular task. All values are in $\%$, and higher is better. Significance $p \leq 0.05$.}}
  \label{tab:supp_diffInternAnnotatorConsensusThresholds}
  \vspace*{-0.3cm}
\end{table*}

\SM{In main, we evaluated our models with an inter-annotator agreement thresholds of 78\%, meaning at least 7 out of 9 agree for each annotation instance 
(`Annotator agreement on best view' in Sec.~\ref{sec:datasets} in main). 
Here, we evaluate even higher agreement thresholds of 89\%---at least 8 out of 9 annotators agree, and 90\%---all annotators agree. 
For HT100M~\cite{miech2019howto100m}, the number of samples drops from 3,151 to 1,840 at 80\%, and 1,345 at 90\%. For Ego-Exo4D~\cite{grauman2023ego}, the same goes down from 5,049 to 3,421 at 80\%, and 1,887 at 90\%, respectively.
Table~\ref{tab:supp_diffInternAnnotatorConsensusThresholds} reports the results. Even at these higher and more challenging inter-annotator agreement thresholds, our model outperforms the strongest baseline---Retrieval~\cite{wang2024internvideo2}-$F$ for view-switch detection, and 
LangView~\cite{majumder2024viewpoint}-bigData
for view selection---on both tasks.}

\subsection{\SM{Finetuning jointly with narration-based pseudo-labels and best view labels, for view selection}}\label{sec:weakSuperv_switchPlusLavis}

\SM{Here, we provide details on how we finetune our view-switch detector jointly with narration-based pseudo-labels~\cite{majumder2024viewpoint} and our best view labels (`Data for view selection with limited labels' in Sec.~\ref{sec:datasets} in main), for doing view selection. Essentially, we modify our view selector training loss $\mathcal{L}^S$ 
(Sec.~\ref{sec:training} in main) 
as follows:
\begin{align}
    \mathcal{L}^S = \mathcal{L}_{\text{CE}} (\ddot{V}_{(t_w, t_w + \Delta]}, V_{(t_w, t_w + \Delta]}) + \alpha *  \mathcal{L}^{N'},
\end{align}
where $\mathcal{L}^{N'} = \mathcal{L}_{\text{CE}} (\ddot{V}_{(t_w, t_w + \Delta]}, \tilde{V}^{N'}_{(t_w, t_w + \Delta]})$ is the cross-entropy loss between the predicted views $\ddot{V}_{(t_w, t_w + \Delta]}$ 
(Sec.~\ref{sec:fewShot_viewSelection} in main) 
and narration-based pseudo-labels~\cite{majumder2024viewpoint} $\tilde{V}^{N'}_{(t_w, t_w + \Delta]}$, generated using next narrations $N'$ 
(Sec.~\ref{sec:pretextTask_def} in main),
and $\alpha$ is the weight on $\mathcal{L}^{N'}$, which we set $\alpha$ to 0.3 on the basis of validation. 
}

\subsection{Duration of past frames}\label{sec:supp_pastFrameDurn}

\begin{table}[!t]
\small
  \centering
  \setlength{\tabcolsep}{4pt}
    \begin{tabular}{l c c c}
    \toprule

     Model & Accuracy & AUC & AP\\
     \midrule
    $T^F = 2$ & \textbf{59.4} & 63.1 & 59.6\\
    $T^F = 4$ & \underline{59.0} & \underline{63.4} & \underline{60.2}\\
    \textbf{$T^F = 8$ (Ours)} & \textbf{59.4} & \textbf{63.8} & \textbf{60.5}\\
    $T^F = 16$ & 55.4 & 59.1 & 57.0\\
    $T^F = 32$ & 52.8 & 55.0 & 53.6\\
    \bottomrule
  \end{tabular}
  \caption{
  Analysis of the impact of the duration of past frames, denoted with $T^F$, on view-switch detection performance on the HowTo100M~\cite{miech2019howto100m} dataset. All values are in $\%$, and higher is better. Significance $p \leq 0.05$.
  }
  \label{tab:supp_pastFrameDurn}
  \vspace*{-0.3cm}
\end{table}

In Table~\ref{tab:supp_pastFrameDurn}, we report our view-switch detection performance numbers for different durations of past frames, denoted by $T^F$, using the HowTo100M~\cite{miech2019howto100m} dataset. We notice that our model performance declines monotonically as we move from our choice of $T^F = 8$ seconds 
\SM{(`Implementation' in Sec.~\ref{sec:experiments} in main)}
to both smaller and larger values. While very short visual contexts fail to capture long-range temporal patterns in human-preferred view sequences, very long visual contexts might contain spurious signals that affect model performance.  $T^F = 8$ seconds balances this trade-off and leads to the best model performance, per this study.

\subsection{Duration of past narrations}\label{sec:supp_pastNrrnDurn}

\begin{table}[!t]
\small
  \centering
  \setlength{\tabcolsep}{4pt}
    \begin{tabular}{l c c c}
    \toprule
     Model & Accuracy & AUC & AP\\
     \midrule
    $T^N = 2$ & 56.1 & 55.9 & 56.2\\
    $T^N = 4$ & 52.4 & 53.9 & 53.4\\
    $T^F = 8$ & 55.5 & 60.2 & 58.0\\
    $T^N = 16$ & \underline{56.1} & \underline{60.2} & \underline{58.0}\\
    \textbf{$T^N = 32$ (Ours)} & \textbf{59.4} & \textbf{63.8} & \textbf{60.5}\\
    \bottomrule
  \end{tabular}
  \caption{
  Analysis of the impact of the duration of past narrations, denoted with $T^N$, on view-switch detection performance on the HowTo100M~\cite{miech2019howto100m} dataset. All values are in $\%$, and higher is better. Significance $p \leq 0.05$.
  }
  \label{tab:supp_pastNrrnDurn}
  \vspace*{-0.3cm}
\end{table}

In Table~\ref{tab:supp_pastNrrnDurn}, we report our view-switch detection results for different durations of past narrations, denoted by $T^N$, with HowTo100M~\cite{miech2019howto100m}. Upon reducing $T^N$ to values lower than our choice of 32 seconds 
\SM{(`Implementation' in Sec.~\ref{sec:experiments} in main),}
our model performance declines monotonically. This shows that a longer past narration context helps better learn correlations between the text in the narrations and desired view types.

\subsection{\SM{Scenario-level analysis of view-selection performance}}\label{sec:sceneLevel_breakdown}

\begin{figure}[b]
    \centering
    \includegraphics[width=0.6\linewidth]{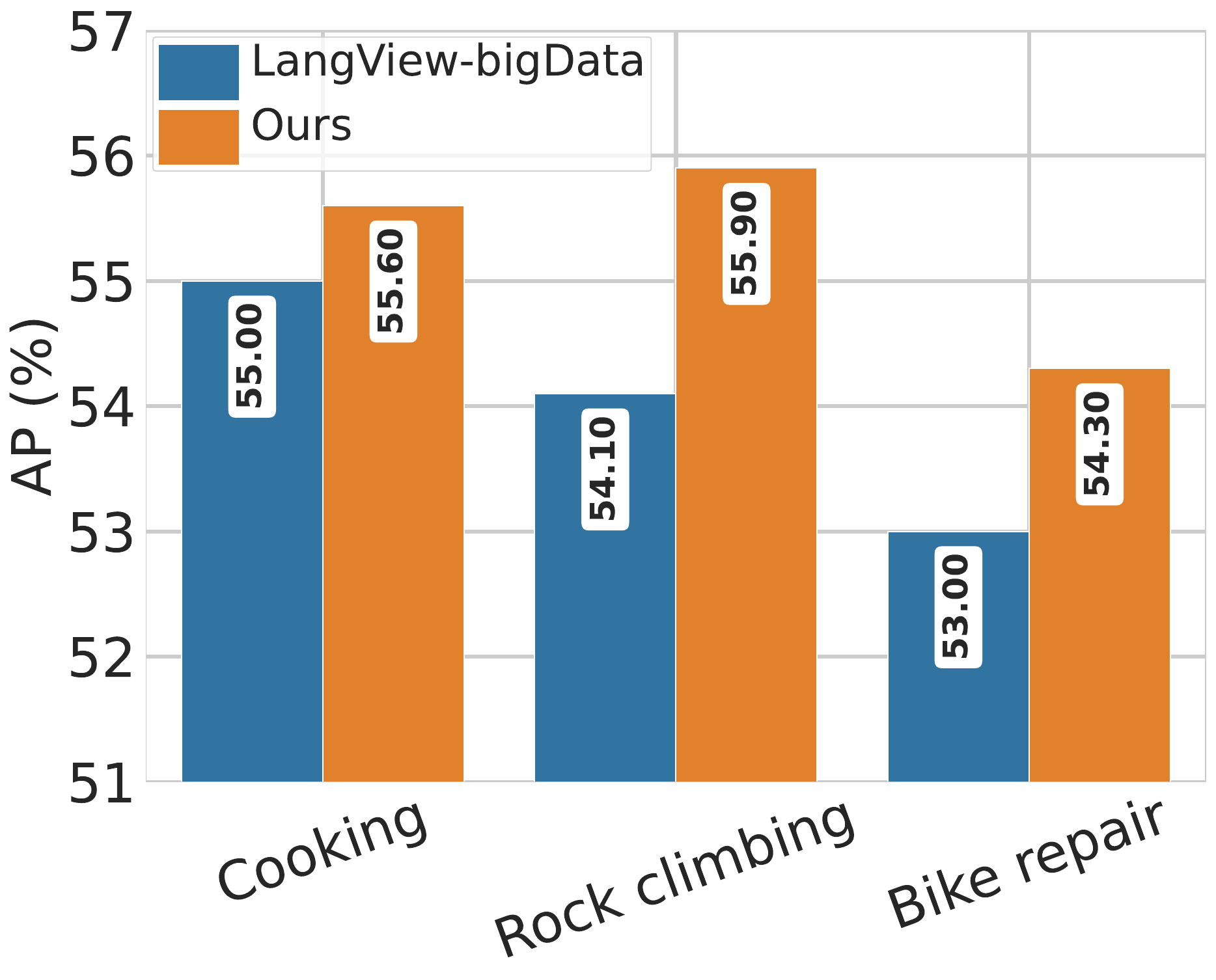}
    \caption{
    \SM{Per-scenario breakdown of our and the strongest baseline, 
    LangView-bigData's
    view selection performance, measured with AP ($\%$).
    }
    }
    \label{fig:sceneLevel_analysis}
\end{figure}

\SM{Fig.~\ref{fig:sceneLevel_analysis} shows the  breakdown of view selection performance per scenario, where only the scenarios with a minimum of 10
instances after filtering low-quality annotations 
(`Data for view selection with limited labels' in Sec.~\ref{sec:datasets} in main) 
are shown. Compared to the best-performing baseline, 
\SM{LangView-bigData},
our model's performance goes up both in absolute and relative terms, from the procedural scenarios like Cooking or Bike Repair, to physical scenarios like Rock climbing. This demonstrates that our model is better able to handle more scenarios with more physical activity, and consequently, more view changes, than the best baseline.
}

\subsection{\SM{Feature similarity baselines with CLIP~\cite{radford2021learning} encoders}}\label{sec:baseline_w_clipEnc}
\begin{table}[!t]
\small
  \centering
  \setlength{\tabcolsep}{4pt}
  \resizebox{\linewidth}{!}{
    \begin{tabular}{l c c c | c c c}
    \toprule
     &  \multicolumn{3}{ c| }{HowTo100M~\citep{miech2019howto100m}} & \multicolumn{3}{ c }{Ego-Exo4D~\citep{grauman2023ego}} \\
     Model & Accuracy & AUC & AP &  Accuracy & AUC & AP\\
    \midrule
    Retrieval~\cite{radford2021learning}-$F$ & \underline{53.0} & \underline{53.0} & 52.7 &  \textbf{52.1} & \underline{52.1} & \underline{53.4}\\
    Retrieval~\cite{radford2021learning}-$N$ & 52.3 & 52.3 & 51.8 &  51.8 & 51.8 & 50.7\\
    Retrieval~\cite{radford2021learning}-$N'$ & 52.6 & 52.6 & \underline{53.2} &  52.0 & 52.0 & 52.5\\
    \textbf{Ours} & \textbf{59.4} & \textbf{63.8} & \textbf{60.5} & \underline{51.2} & \textbf{56.4} & \textbf{55.4}\\
    \bottomrule
  \end{tabular}
  }
  \caption{
  View-switch detection results. 
  Evaluation on Ego-Exo4D~\cite{grauman2023ego} is zero-shot. All values are in $\%$, and higher is better. 
  Significance $p \leq 0.05$.
  }
  \label{tab:vsd_simBaselinesWclip}
  \vspace*{-0.2cm}
\end{table}

\begin{table}[!t]
\small
  \centering
  \setlength{\tabcolsep}{4pt}
    \begin{tabular}{l c c c }
    \toprule
     Model & Accuracy & AUC & AP\\
    \midrule
    Retrieval~\cite{radford2021learning}-$F$ & 52.1 & 52.1 & \underline{53.4}\\
   Retrieval~\cite{radford2021learning}-$N$ & 51.8 & 51.8 & 50.7\\
    Retrieval~\cite{radford2021learning}-$N'$ & 52.0 & 52.0 & 52.5\\
    View-narration~\cite{ma2022x} Similarity  & \underline{52.5} & \underline{52.2} & \underline{53.4}\\
    \textbf{Ours} & \textbf{54.0} & \textbf{57.3} & \textbf{56.0}\\

    \bottomrule
  \end{tabular}
  \caption{
  \SM{Results and ablation study for view selection with limited
labels.}
  All values are in $\%$, and higher is better. 
  Significance $p \leq 0.05$.
  }
  \label{tab:vs_simBaselinesWclip} 
  \vspace*{-0.3cm}
\end{table}

\SM{
In `View-switch detection' and `View selection' in Sec.~\ref{sec:experiments} in main,
we evaluated feature similarity baselines with InternVideo2~\cite{wang2024internvideo2} encoders. Here, we provide the parallel experiment with CLIP~\cite{radford2021learning}-style encoders in Table~\ref{tab:vsd_simBaselinesWclip} for view-switch detection, and Table~\ref{tab:vs_simBaselinesWclip} for view selection. Specifically, while for the retrieval baselines, we use the unmodified CLIP encoders, we use X-CLIP~\cite{ma2022x} encoders in the View-narration Similarity baseline for encoding multiple frames in the ego and exo views 
(Sec.~\ref{sec:fewShot_viewSelection} in main). 
With CLIP, the similarity baselines generally perform worse. This happens possibly because, unlike InternVideo2, CLIP features are not very fine-grained and/or do not capture temporal context
in the case of retrieval baselines. Furthermore, our model outperforms all feature similarity baselines, even when implemented with CLIP, highlighting its advantages over the baselines across different encoder choices.}

\subsection{Additional dataset details}\label{sec:supp_dataset}
Here, we give further dataset details, in addition to the ones provided in 
\SM{`Training data', `Evaluation data' and `Data for view selection with limited labels' in Sec.~\ref{sec:datasets} in main}.

\paragraph{Charades-Ego~\cite{sigurdsson2018actor} datasets for training view classifier in our pseudo-labeler.}
As mentioned in 
\SM{Sec.~\ref{sec:pseudo_labeler} in main}
in main, we train the view classifier of our pseudo-labeler on the Charades-Ego~\cite{sigurdsson2018charades} dataset. To do so, we create a dataset containing 5,551 train, 615 val, and 1,597 test videos, where all videos are randomly sampled and the splits are completely disjoint. Moreover, we train and test our model by sampling fixed-length clips, where the clip length is set to 2 seconds. 

\paragraph{HowTo100M~\cite{miech2019howto100m} datasets for view-switch detection.}
As noted in 
\SM{`Training data' and `Evaluation data' in Sec.~\ref{sec:datasets} in main}
in main, we train our view-switch detector on HowTo100M (HT100M)~\cite{miech2019howto100m} and also use it as a dataset for evaluating view-switch detection. To do so, we sample a maximum of 500 videos from each category in the second level of the HT100M video classification hierarchy. This results in a total of 4\SM{,}391 hours of HT100M videos.

In addition to the details provided in 
\SM{`Evaluation data' in Sec.~\ref{sec:datasets} in main},
for creating the HT100M test set for evaluating view-switch detection, 
we 
also include
the clips right after all view-switch boundaries, as identified by our pseudo-labeler, 
in the test set. This ensures that the test set is not totally dominated by the more frequently-occurring same-view instances, which can affect the estimation of our unbiased mean performance 
\SM{(`Evaluation metrics' in Sec.~\ref{sec:experiments} in main)}.
\SM{Finally, we provide the clip just before each clip being labeled, to the annotators
in order to identify instances where the \emph{ground-truth} view stays the same (same-view) and where it switches (view-switch). This allows us to separately evaluate
these two alternate but important scenarios, and report \emph{unbiased} mean performance (`Evaluation metrics' in Sec.~\ref{sec:experiments} in main).}

\paragraph{Ego-Exo4D~\cite{grauman2023ego} datasets.}
We create our datasets for Ego-Exo4D~\cite{grauman2023ego} 
\SM{(`Evaluation data' and `Data for view selection with limited labels' in Sec.~\ref{sec:datasets} in main)}
for evaluating view-switch detection, and training and evaluating view selection, by sampling clips from each video at a regular interval of 1 second.

\subsection{Additional annotation details}\label{sec:supp_annotation}
Here, we provide further annotation details, in addition to what are provided in
\SM{(`Evaluation data' and `Annotator agreement on best view' in Sec.~\ref{sec:datasets} in main)}.
We start with the details that are common for both 
\SM{both HT100M~\cite{miech2019howto100m} and Ego-Exo4D~\cite{grauman2023ego}.}

We use Amazon Mechanical Turk (MTurk) to collect annotations for both
\SM{datasets}.
\SM{Before assigning an MTurk worker our job, we ensure that their prior annotation approval rate is more than $98\%$. We also require them to take pass short qualifiers 
(`Annotator agreement on best view' in Sec.~\ref{sec:datasets} in main),
each of which contains 10 annotation instances. We design these qualifiers such that they are very similar to our main jobs. Furthermore, we handpick the annotation instances in the qualifiers such we that know the ground-truth for each of them. This lets us easily compare an annotator's choices against the ground-truths in the qualifiers, and consequently, gauge the their reliability. We only accept annotators who pass these qualifiers with a success rate of at least 90\%.}

Next, we provide 
\SM{dataset}-specific annotation details.

\paragraph{\SM{HT100M}.}

\begin{figure*}[t]
    \centering
    \includegraphics[width=1.0\linewidth]{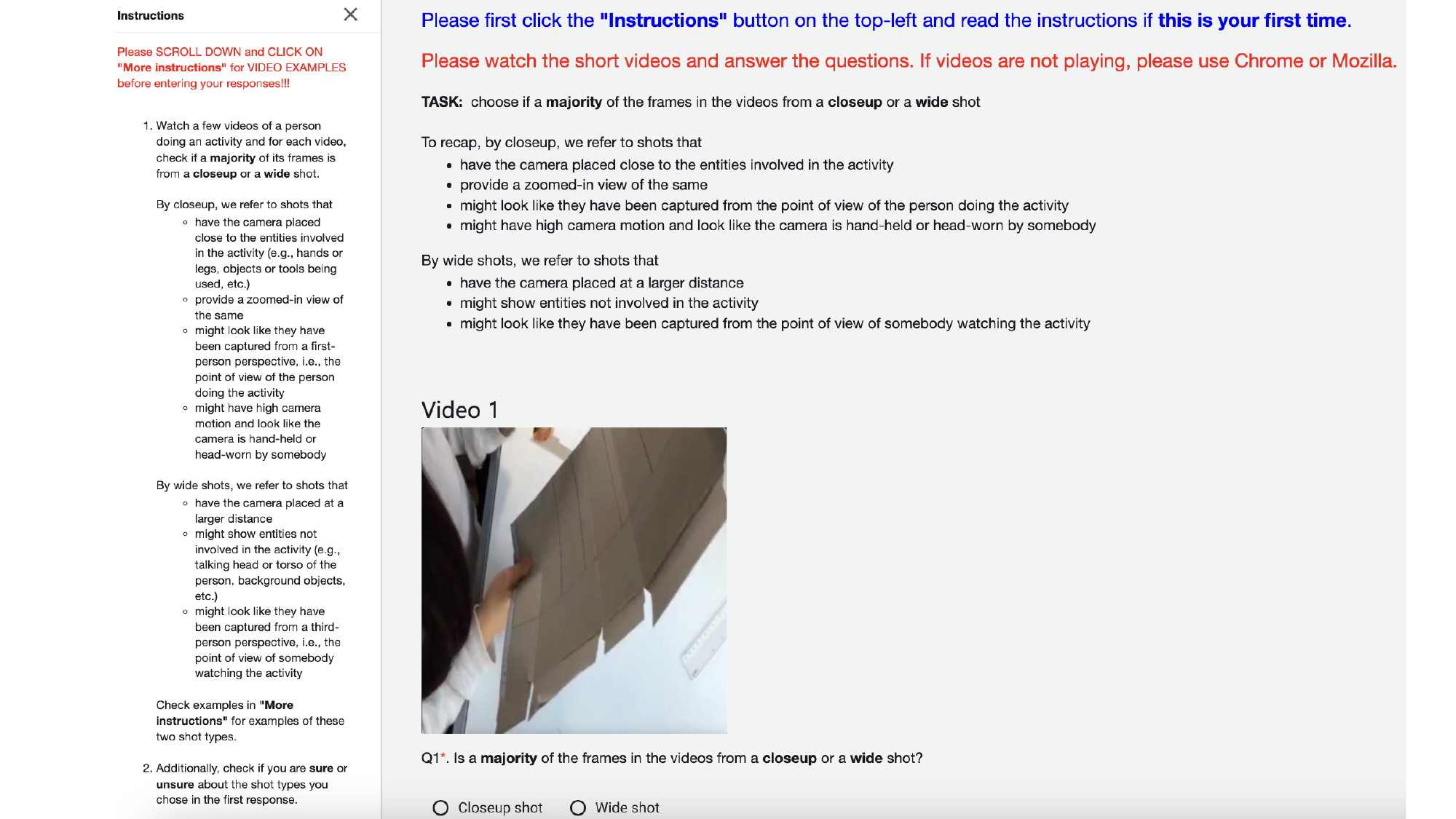}
\caption{Sample interface for collecting 
\SM{HT100M~\cite{miech2019howto100m} annotations 
(`Evaluation data' and `Annotator agreement on best view' in Sec.~\ref{sec:datasets} in main)}.
Additionally, we also provide video examples for both ego (closeup shot) and exo (wide shot) clips, to help the annotators.}
\label{fig:supp_annotationInterface_ht100m}
\end{figure*}

In Fig.~\ref{fig:supp_annotationInterface_ht100m}, we show our annotation interface for 
\SM{HT100M (`Evaluation data' and `Annotator agreement on best view' in Sec.~\ref{sec:datasets} in main)}.
In short, we provide a set of detailed instructions, which lists the different characteristics of both ego and exo clips\footnote{\SM{We refer to ego and exo views as ``closeup" and ``wide" shots, respectively, in order to easily explain the annotation process to the workers.}}, and give examples for each characteristic. 
Additionally, we provide a more concise summary of the lists of per-view attributes on each annotation page to give a quick recap of the annotation task, to the workers. Finally, we provide video examples of both ego and exo clips, to further guide the annotation process.

\paragraph{\SM{Ego-Exo4D}.}

\begin{figure*}[t]
    \centering
    \includegraphics[width=0.85\linewidth]{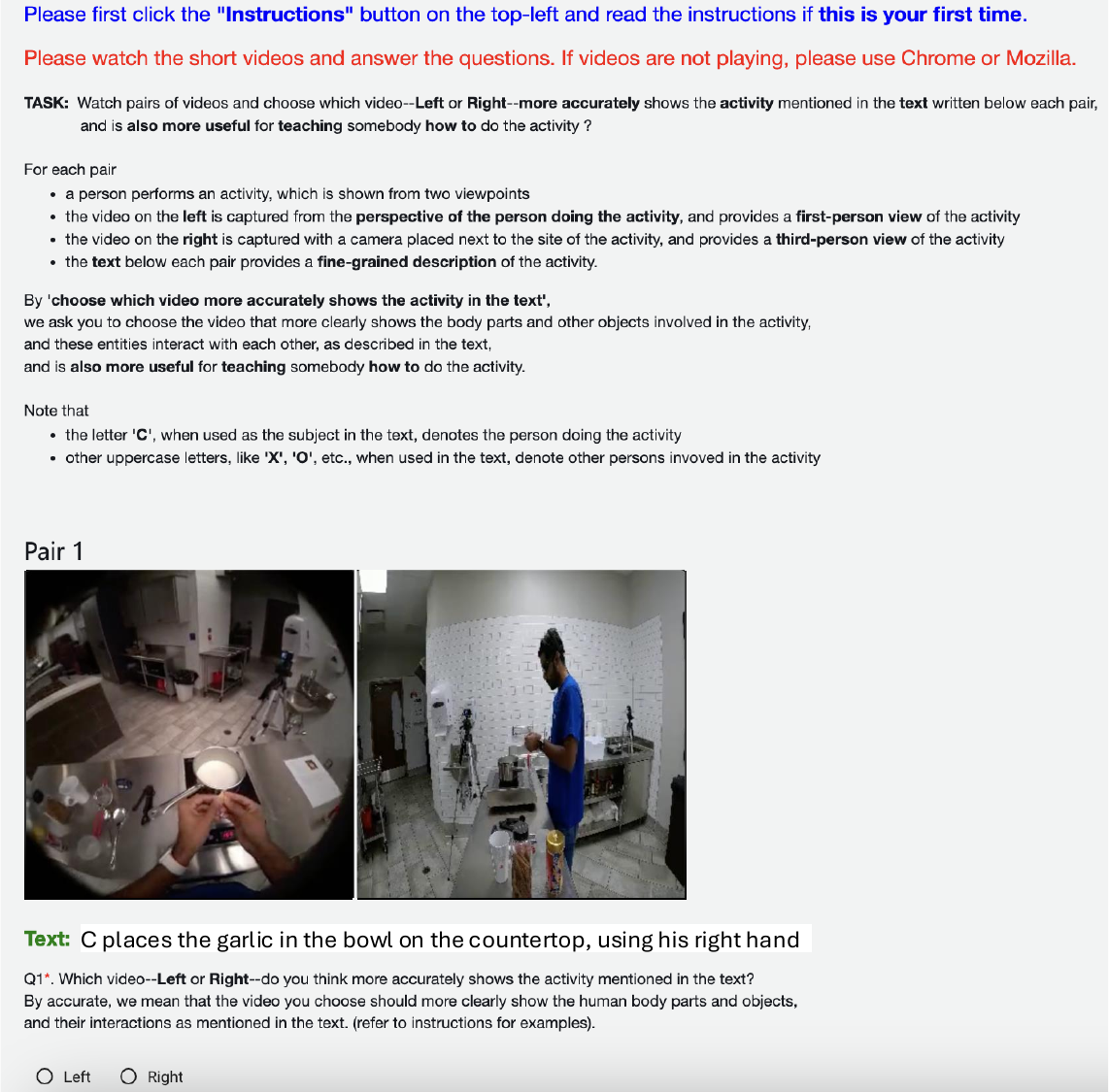}
\caption{Sample interface for collecting 
\SM{Ego-Exo4D~\cite{grauman2023ego} annotation 
(`Evaluation data' and `Annotator agreement on best view' in Sec.~\ref{sec:datasets} in main)
}.
Additionally, we also provide examples showing pairs of clips from both ego and exo views, their associated atomic descriptions, and how to reason about which view is better for viewing an activity, to help the annotators.}
\label{fig:supp_annotationInterface_egoExo}
\end{figure*}

In Fig.~\ref{fig:supp_annotationInterface_egoExo}, we show our annotation interface for
\SM{Ego-Exo4D (`Evaluation data' and `Annotator agreement on best view' in Sec.~\ref{sec:datasets} in main)}.
In summary, we provide a set of detailed instructions on each annotation page, which describes the kind of information captured by the two views (ego and exo) and the role of the associated atomic description, and also specify that we expect the annotator to pick a view that best shows the activity mentioned in the atomic description and hence, useful for instructional purposes. To further assist with the annotation process, we provide examples showing pairs of clips from both ego and exo views, their associated atomic descriptions, and how an annotator should reason about which view is better for viewing the activity, in the context of its corresponding atomic description.

\subsection{Additional implementation details}\label{sec:supp_implementation}
Here, we provide additional implementation details.

\paragraph{View assignment to past frames and narrations for Ego-Exo4D~\cite{grauman2023ego}.}
In 
\SM{Sec.~\ref{sec:pseudo_labeler} in main},
we provide details about how we assign our view pseudo-labels to past frames and past narrations, for using our model (view-swtich detector or view selector) with HowTo100M~\cite{miech2019howto100m}. Here, we describe our process for assigning views to the past frames and narrations for Ego-Exo4D~\cite{grauman2023ego}.

Note that all frames or narrations in an Ego-Exo4D video might not be assigned a ground-truth best view during our annotation process, because the annotations for some couplets of pairs of clips from the two views, and their associated atomic descriptions 
\SM{(`Evaluation data' in Sec.~\ref{sec:datasets} in main)},
might get discarded due to our annotation quality control measures 
\SM{(`Annotator agreement on best view' in Sec.~\ref{sec:datasets} in main, and above)}.
To tackle this, we set the best view for each past frame and past narration to the best view ground-truth for the nearest couplet of clip pair and its associated atomic description, for which the ground-truth has not been discarded.

\subsubsection{View pseudo-labeler}
\paragraph{Scene detector.}
As mentioned in 
\SM{Sec.~\ref{sec:pseudo_labeler} in main},
we use PySceneDetect~\cite{pyscenedetect}, an off-the-shelf scene detector, for detecing scene boundaries in HowTo100M~\cite{miech2019howto100m} videos. Specifically, we use the image-content-based detector. Moreover, we set the weights for pixel colors to 1.0 and the same for object edges to 0.0, and the minimum shot length to 2 seconds.  
 
\paragraph{View classifier.}
For our view classifier 
\SM{(Sec.~\ref{sec:pseudo_labeler} in main)}, 
we use the slow branch of a SlowFast~\cite{feichtenhofer2019slowfast} model that has a ResNet3D~\cite{tran2018closer}-50 architecture and is pretrained on the Kinteics-400~\cite{kay2017kinetics} dataset, as the visual encoder. This encoder takes 8 uniformly sampled frames from each 2-second clip (c.f.~Sec.~\ref{sec:supp_annotation}), embeds them into a visual feature, and passes the visual feature to a linear classification head, which is implemented as a single linear layer. We initialize the parameters for all model components that are trained from scratch, using a parameter initialization strategy for masked auto-encoders for videos~\cite{feichtenhofer2022masked}. We train the model until convergence by using an AdamW~\cite{loshchilov2017decoupled} optimizer, a batch size of 32, and initial learning rates of $10^{-5}$ and $10^{-4}$ for the visual encoder, and the classification head, respectively. 

\subsubsection{View-switch detector}
Here, we provide more implementation details of view-switch detector, in addition to those provided in 
\SM{`Implementation' in Sec.~\ref{sec:experiments}}
in main. Our detector's DINOv2~\cite{oquab2023dinov2} frame encoder has 12 layers and takes in frames sampled at 4 fps, and produces a 768-dimensional feature for each frame. Our detector's Llama 2~\cite{touvron2023llama2} text encoder begins by producing a token sequence for each input narration by tokenizing its text, and padding the tokenizer output to match the length of the longest token sequence in a batch, or truncating it to reduce its length to 512, depending on which length is shorter. Moreover, for the past narrations, it ensures that their total token count does not cross 1024, by truncating wherever necessary. It then encodes each token from the past narration sequence, 
or the \SM{next} narration, 
into a 4096-dimensional features. Next, it projects each such feature into  a 768-dimensional feature using a linear layer. We implement our modules for encoding views for both frames and past narrations, and \SM{their} relative temporal positions,
as learnable 
embeddings 
that produce 768-dimensional features. \SM{Specifically, for encoding views, we use a learnable feature dictionary with 0 (ego) and 1 (exo) as keys. For encoding relative temporal positions, we first discretize the durations in seconds, by using bins of size 0.1 second, and then encode them with learnable feature dictionaries, in which the number of keys is set to the maximum number of bins.}
To aggregate all the above-mentioned 768-dimensional features, we use a 8-layer and 8-head transformer encoder~\cite{vaswani2017attention} that adds a positional embedding~\cite{vaswani2017attention} to each feature and performs self-attention on them. Finally, the 2-layer MLP for our view classification is a stack of two hidden linear layers with the output feature size of 256 and 64, respectively, and a final linear layer that esimtates the next view. We initialize the parameters for all model components that are trained from scratch, using a parameter initialization strategy for masked auto-encoders for videos~\cite{feichtenhofer2022masked}. We freeze the pre-trained components of our view-switch detector and train the rest of the model until convergence with the AdamW~\cite{loshchilov2017decoupled} optimizer, a batch size of 48, and a learning rate of $10^{-6}$.

\subsubsection{View selector}
Our view selector has the exact same architecture as our view-switch detector. For training it, we first initialize its parameters with those of our pretrained view-switch detector. We then freeze the frame encoder and the text encoder, and finetune the rest of the model until convergence with the exact same optimizer and hyperparameters as the ones used in our view-switch detector training. 